\documentclass[letterpaper, 10 pt, conference]{ieeeconf}  

\IEEEoverridecommandlockouts                              

\title{\LARGE \bf
Real-time Two-tape Control System in Vine robots
}

\author{Hanmo Liu, Kayleen Smith, Zimu Yang, Mark Yim}
\usepackage{algorithm}
\usepackage{float}
\usepackage{algpseudocode}
\usepackage{amsmath}
\usepackage{graphicx}
\usepackage{xcolor}
\usepackage{caption}
\let\labelindent\relax
\usepackage{enumitem}
\usepackage{caption}
\usepackage{subcaption}
\usepackage{afterpage}
\usepackage{url}

\begin{document}

\maketitle
\thispagestyle{empty}
\pagestyle{empty}

\begin{abstract}

This paper focuses on how to make a growing Vine robot steer in different directions with a novel approach to real-time steering control by autonomously applying adhesive tape to induce a surface wrinkles. This enabling real-time directional control  with arbitrary many turns while maintaining the robot's soft structure. This system feeds growing material external to the tube. The design achieves fixed-angle turns in 2D space. Through experimental validation, we demonstrate repeated 21-degree turns using a Dubins path planner with minimal error, establishing a foundation for more versatile Vine robot applications. This approach  combines real-time control, multi-degree-of-freedom steering, and structural flexibility, addressing key challenges in soft robotics.

\end{abstract}

\section{INTRODUCTION}

Vine robots are innovative soft robots that mimic plant vine growth \cite{kiryu2008grow, blumenschein2020design,del2024growing}. These robots offer several unique advantages: they can safely handle heavy loads using high-pressure gas, navigate through confined spaces due to their flexible tubular structure \cite{stroppa2020human}, theoretically extend up to material supply \cite{blumenschein2020design}, and execute multiple turns in any direction. These capabilities make them particularly useful for applications in search and rescue \cite{zhou2024development} and exploration of narrow spaces \cite{8917931}, such as archaeological tunnels \cite{zhou2024development, girerd2024material}, or 3D electromagnetic reconfiguration \cite{gan20203d}. 

However, steering control remains a significant challenge in Vine robot development. Previous approaches have struggled to achieve three objectives: real-time steering, multi-degree-of-freedom (multiDOF) control, and maintenance of a soft structure. Existing solutions include preformed shapes where the path is predetermined \cite{hawkes2017soft} (not real-time), artificial muscle-controlled designs \cite{gan20203d} or internal rigid  bend actuators \cite{haggerty2021hybrid} (not multiDOF), and 3D-printed variants \cite{del2024growing} (not soft).

To address these limitations, we propose a novel approach to steering control using real-time, induced surface wrinkles. Our design uses an exterior material feed and an innovative end-device system that enables dynamic wrinkle generation and fixation using a removeable adhesive tape.   A heat sealing method for steering has been previously shown, similar to our approach, in which bends are introduced at the head of the robot, but the bends in this case are permanent \cite{satake2020novel}.

After introducing the approach and design, we also include a mathematical model showing consistent $21^o$ turns and a modified path planning algorithm using this fixed angle constraint. We then show the results physical experiments and conclude with limitations and future work. 

\section{RELATED WORK}

Preformed methods have been a foundational approach in Vine robot steering control. This technique places adhesive tape on the robot's surface before deployment so the path forms during deployment. The fundamental principle relies on creating asymmetric surface constraints. As the robot grows, the tape restricts expansion on one side while allowing normal growth on the other, resulting in controlled bending motions \cite{hawkes2017soft}.

A mathematical model that quantifies the correlation between tape configurations and steering angles, enabling more precise control over the robot's growth trajectory was also developed \cite{blumenschein2020design}. However, despite these advances, pre-taped methods cannot deviate from this preset trajectory to adapt to unexpected obstacles or changing environmental conditions limiting real-world applications.
\section{SYSTEM DESIGN}

\begin{figure}[h]
    \begin{subfigure}[b]{0.47\textwidth}
        \includegraphics[width=\textwidth,angle=0]{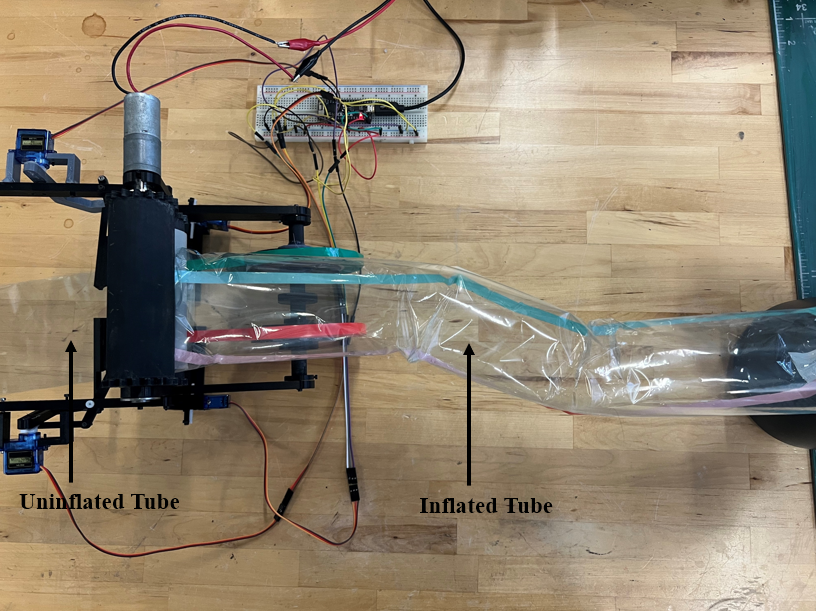}
        \caption{The end-device is shown on the left inflated tube in the middle and the air pump is on the right (not shown)}
        \label{fig:Overview}
    \end{subfigure}

    \begin{subfigure}[b]{0.47\textwidth}
        \centering
        \includegraphics[width=\textwidth,angle=0]{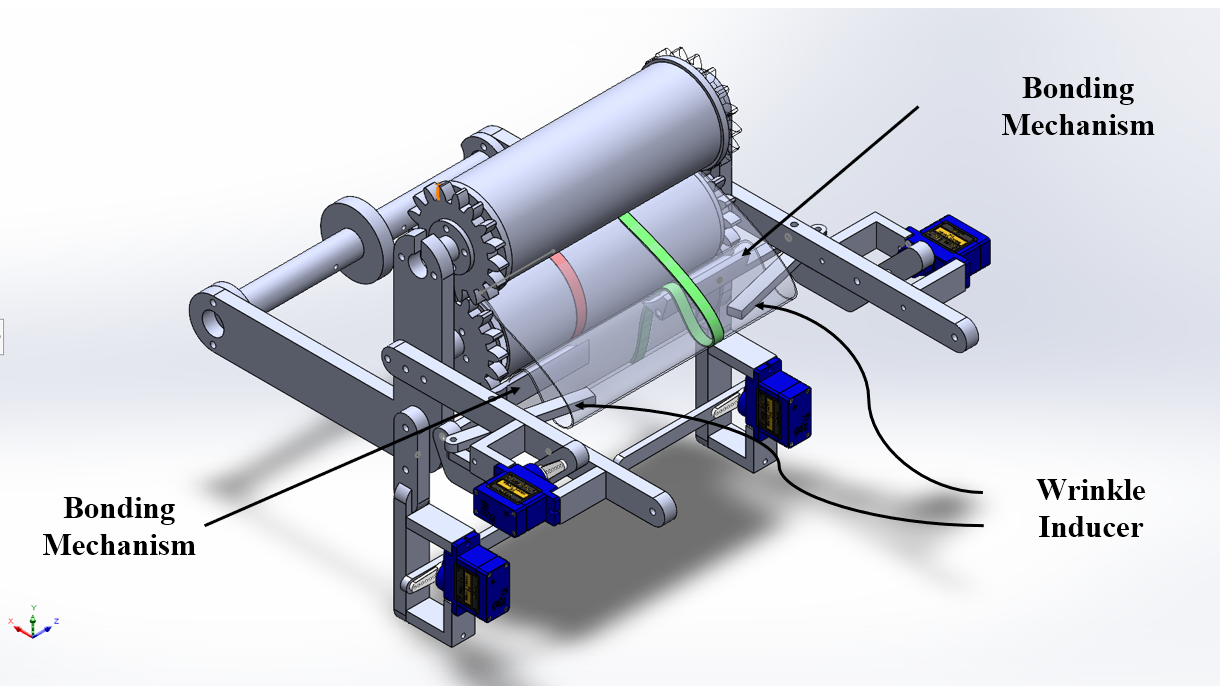}
        \caption{Wrinkle Inducer working position has the wrinkle formed and the green tape following the wrinkle. }
    \end{subfigure}
    \caption{ (a) Overview of robot. (b) CAD model of end-device.}
    \label{fig:CAD}
\end{figure}

\begin{figure*}[h]
    \centering
    \begin{subfigure}[b]{0.25\textwidth}
        \centering
        \includegraphics[width=\textwidth,height = 0.13\textheight]{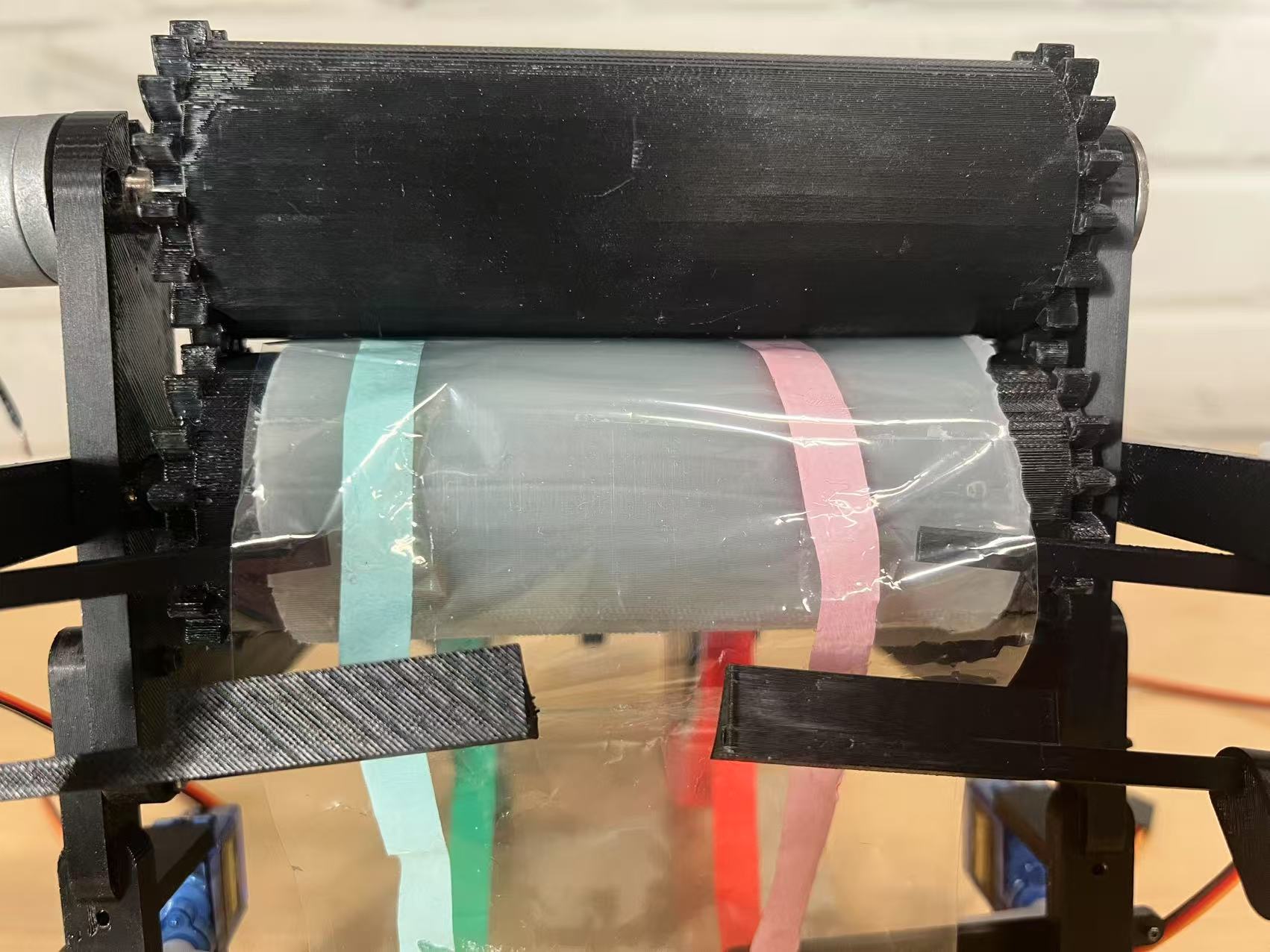}
        \caption{Front view no wrinkle}
        \label{fig:Front View1}
    \end{subfigure}%
    \hfill%
    \begin{subfigure}[b]{0.25\textwidth}
        \centering
        \includegraphics[width=\textwidth,height = 0.13\textheight]{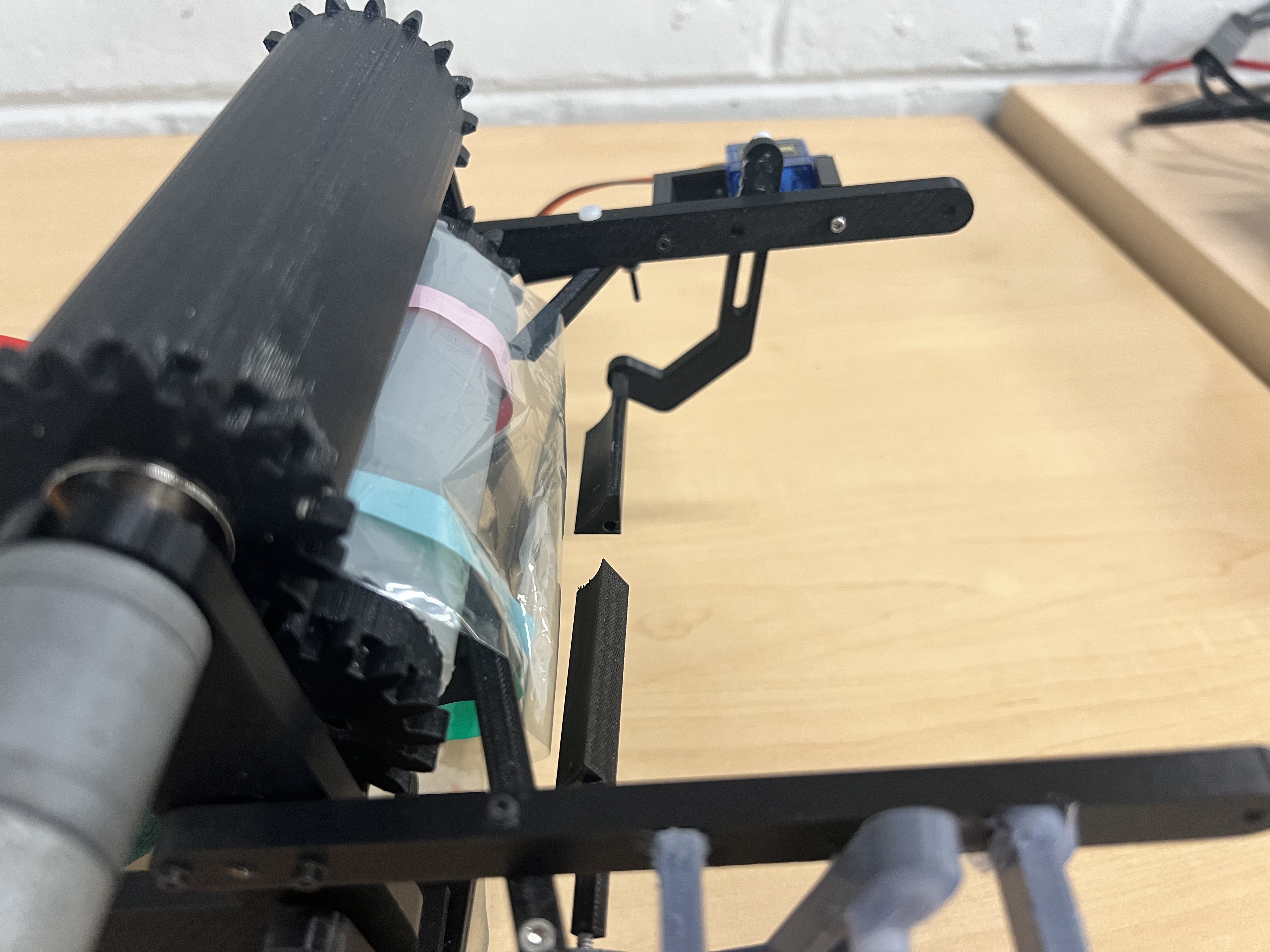}
        \caption{Side view no wrinkle}
        \label{fig:Side View1}
    \end{subfigure}%
    \hfill%
    \begin{subfigure}[b]{0.25\textwidth}
        \centering
        \includegraphics[width=\textwidth,height = 0.13\textheight]{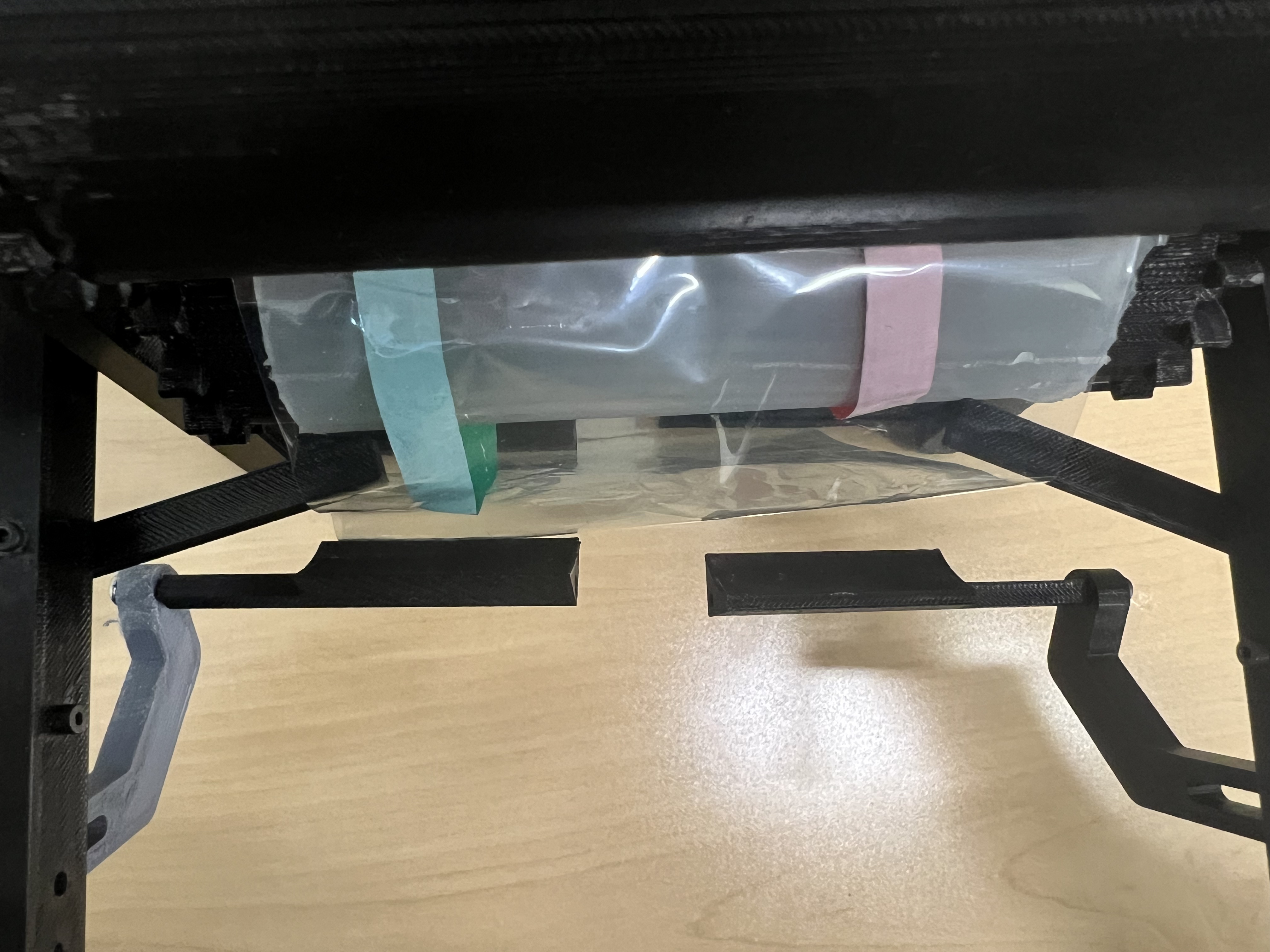}
        \caption{Top view no wrinkle}
        \label{fig:Above View1}
    \end{subfigure}
    \vspace{0.3cm}
    \begin{subfigure}[b]{0.25\textwidth}
        \centering
        \includegraphics[width=\textwidth,height = 0.13\textheight]{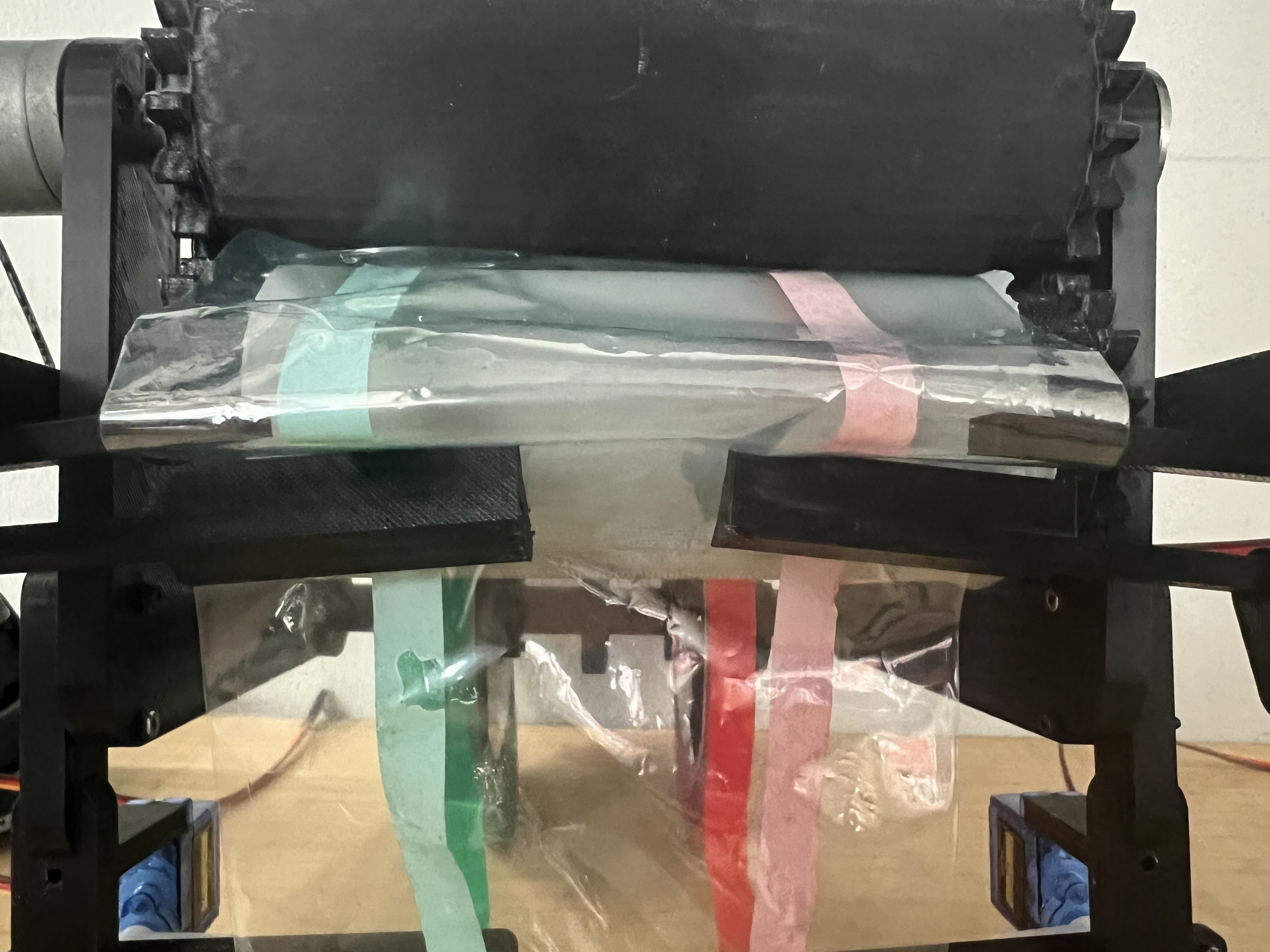}
        \caption{Front view with wrinkle}
        \label{fig:Front View2}
    \end{subfigure}%
    \hfill%
    \begin{subfigure}[b]{0.25\textwidth}
        \centering
        \includegraphics[width=\textwidth,height = 0.13\textheight]{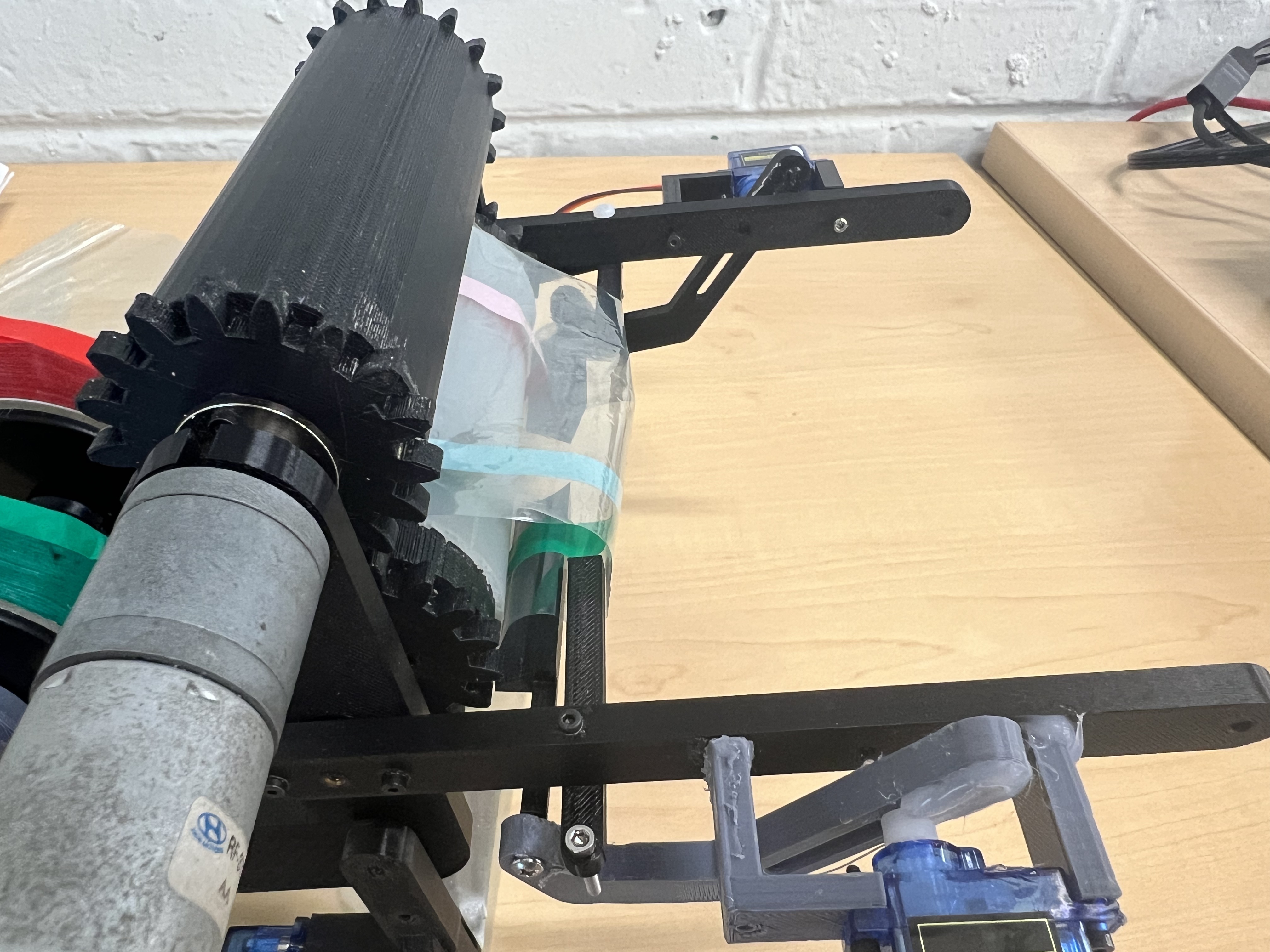}
        \caption{Side view with wrinkle}
        \label{fig:Side View2}
    \end{subfigure}%
    \hfill%
    \begin{subfigure}[b]{0.25\textwidth}
        \centering
        \includegraphics[width=\textwidth,height = 0.13\textheight]{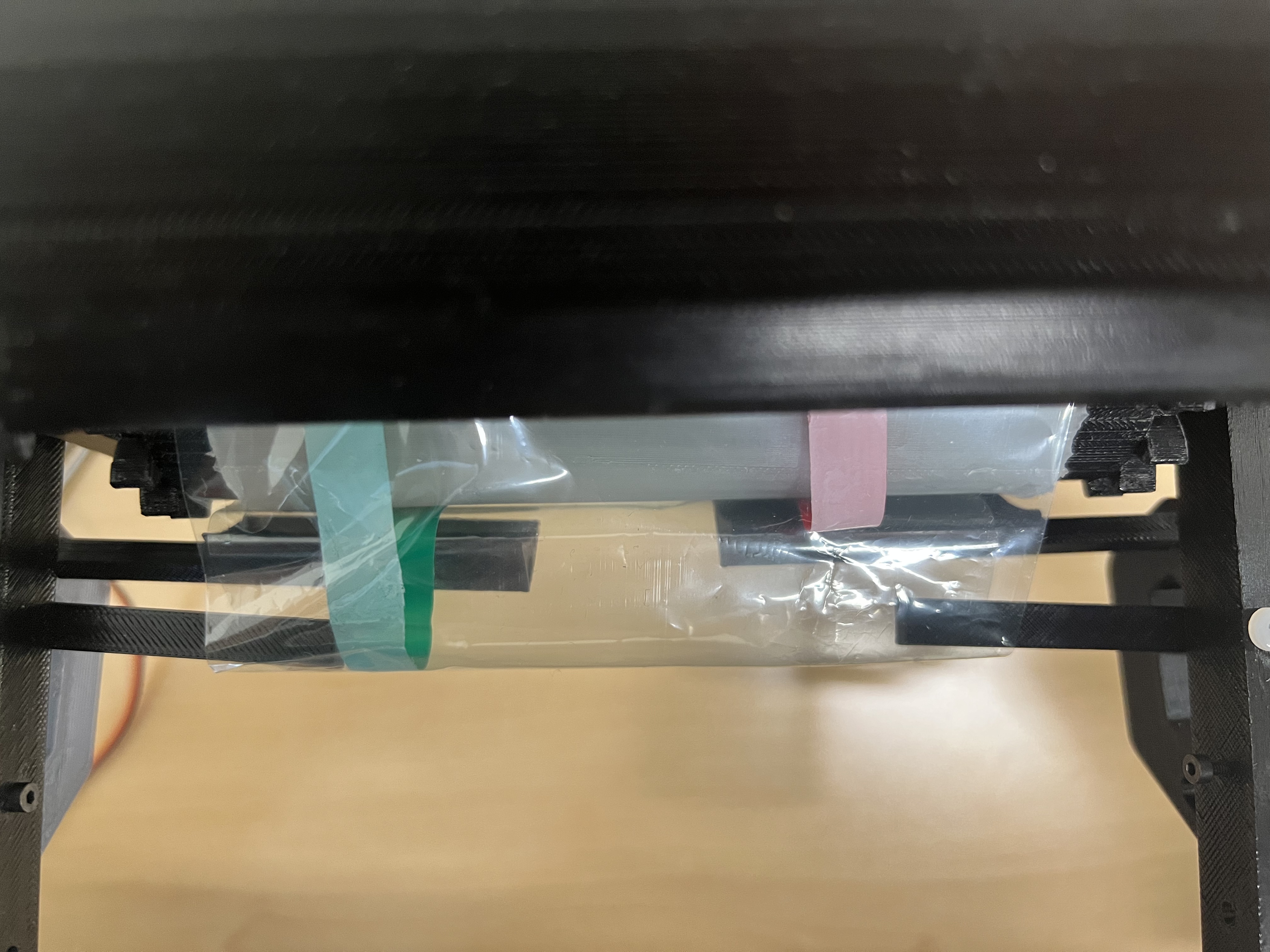}
        \caption{Above view with wrinkle}
        \label{fig:Above View2}
    \end{subfigure}
    \caption{(a)-(c): Wrinkle Inducer in a neutral configuration prior to activation
(d)-(f): Two arms on each side of the wrinkle inducer are used to move the material out and create a fold.}
    \label{fig: Process of Making a Wrinkle}
\end{figure*}

Like other growing Vine robots \cite{blumenschein2020design, greer2019soft}, we use  plastic tubing, used for packaging, laid flat as the robot's main body then inflate that tube to extend the robot. Our tube has a width of 105 mm and is made of PVC.
We use the same method of forming turns as the pre-taped method, but do this on the fly by inducing and taping wrinkles at the growing end of the robot.

Unlike many Vine robots \cite{greer2019soft}, which use an everting method of feeding material inside the growing tube, we place the feeding material (flattened and rolled) outside the inflated part. In this way, wrinkles can be induced into the uninflated roll with an automatic mechanism without having to struggle with the inflated tube pressure. Also, two ribbons of tape run along the full length of the robot as it grows. This ribbon lies flat on the straight parts and does not impede motion; if anything, it strengthens the tube. Hence, when attaching tape to the tube to generate a wrinkle, the length on one side of the tube is shortened relative to the opposite side, creating an asymmetric constraint that forces the tube to bend when inflated. This asymmetric configuration allows the robot to make consistent and controlled turns as it grows.

The end device, as shown in Fig.~\ref{fig:CAD}, which forms the growth point of the system, comprises the following functional components. Two strips of tape are positioned with their non-adhesive sides facing downward and adhesive sides facing upward on the bottom rollers. The PVC tube passes through the gap between the rollers, transitioning from the uninflated to the inflated section. This configuration creates two faces of the tube: an inner face that contacts the adhesive surface of the tape and the rollers and an outer face. The arms of the wrinkle inducer are positioned between the inner face of the tube and the tape. The bonding arm is placed beneath the outer face to control the adhesion between the tube and the tape. During wrinkle formation, the two wrinkle inducer arms rotate outward to create the wrinkle while the bonding arm compresses inward to secure the wrinkle in place.

The parts of the Vine robot, the CAD files can be found online \cite{liu2025data} and are listed here:

\begin{itemize}

\item A source roll of material

\item  An end-device that contains:
\begin{itemize}
    \item Two side-by-side rollers control the growth and squeeze the tube to separate the inflated and uninflated tube sections

    \item A wrinkle inducer  that creates left or right wrinkles.

    \item A lower bonding arm that fixes these wrinkles in place with adhesive tape
\end{itemize}
\item An air pump connected to one end of the  tube  maintains inflated pressure
\end{itemize}
\subsection{Mechanical Design} 

\subsubsection{Wrinkle Inducer and Bonding Arm}
 
The wrinkle inducer serves as the primary steering mechanism for the Vine robot. It consists of two support arms on the left and right sides and two bonding arms underneath. During wrinkle formation, the support arms lift the uninflated portion of the tube off the roller surface while the lower bonding arms compress inward to bond the tube to the adhesive tape. This process creates an asymmetric wrinkle on one side of the tube, which causes the robot to turn in that direction when inflated.

After wrinkle creation, the support arms retract, allowing the newly formed wrinkle to pass smoothly through the roller system to be inflated on the other side. The support arms and bonding arms are synchronized but move in opposite directions and can be driven by one servo.

The length of the wrinkle fold is determined by the length of the support arms in the wrinkle inducer mechanism. Using longer support arms or bigger turning angles create larger wrinkles, corresponding to sharper turning angles when inflating the tube. Our implementation maintains a fixed arm turning angle to simplify the end effector design. This design choice allows the wrinkle inducer and bonding mechanism to be controlled by one motor, streamlining the control system while providing consistent directional control in both left and right directions.
\subsubsection{Rollers}

The roller system is the Vine robot's growth control and air-sealing component. It consists of two adjacent 3D-printed rollers with an intermediate rubber layer. A DC motor drives these rollers, with side-mounted gears to prevent relative slippage between them.
Maintaining an air-tight seal has one main challenges. The rollers must allow the thicker wrinkles to pass through the roller gap, while at all times the seal at the roller interface must prevent pressure leaks from the inflated to the uninflated sections. To achieve this,  a rubber layer that provides elasticity between the rollers effectively accommodating thickness variations while maintaining an air-tight seal.

\subsubsection{Compensation Mechanism}

\begin{figure}[h]
    \centering
    \includegraphics[width=0.3\textwidth,angle=0]{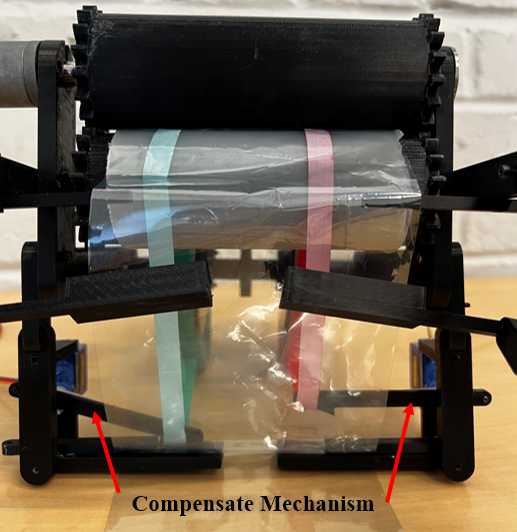}
    \caption{Compensation Mechanism}
    \label{fig:Compensate Mechanism}
\end{figure}

The compensation mechanism  enables asymmetric wrinkles by selectively adhering tape to the opposite side of the tube. Without this mechanism, wrinkle formation would create symmetric folds on both sides, neutralizing the desired turning effect. When creating a directional turn, the compensation mechanism first activates on the non-wrinkle side (e.g., GREEN side). A dedicated control arm presses down to bond the tape directly to the tube surface along its natural path before any wrinkle is formed. As shown in Fig.~\ref{fig:Compensate Mechanism}, when planning to create a wrinkle on the RED side, the compensation arm first attaches the GREEN tape firmly to the tube. This strategic pre-attachment ensures that during the subsequent wrinkle formation phase, only the RED side develops an effective wrinkle while the GREEN side's material follows the natural tube contour without shortening. The timing of this compensation action is crucial --- it must occur before the wrinkle inducer activates to ensure proper asymmetric deformation that produces the desired turning angle when inflated.

\subsection{Workflow of Generating Wrinkles}
Fig.~\ref{fig: Process of Making a Wrinkle}, shows a wrinkle formation on the RED side. The process for forming the wrinkle follows.

\begin{enumerate}[label=Step \arabic*:, leftmargin=*]

\item  Compensation mechanism: The compensation mechanism attaches the tape to the tube on the side with the GREEN tape.

\item  Wrinkle inducer: The wrinkle inducer moves its two arms to create the wrinkle, with the bonding arm attaching the RED tape to the tube by taping the loop of material that forms the wrinkle.  On the GREEN side, the loop of material is also there but is temporary as the GREEN tape has been pre-adhered to follow the inside of the loop and thus unfolds that side of the tube as it inflates.

\item  Roller: After the wrinkle is induced, the two arms return to the standby position, and the roller pushes the wrinkle to the inflated side. The tube inflates and the wrinkle on the RED side makes the Vine robot bend at that wrinkle.

\end{enumerate}

\section{MATHEMATICAL MODELING}

\begin{figure}[t] 
    \centering
    \includegraphics[width=0.5\textwidth]{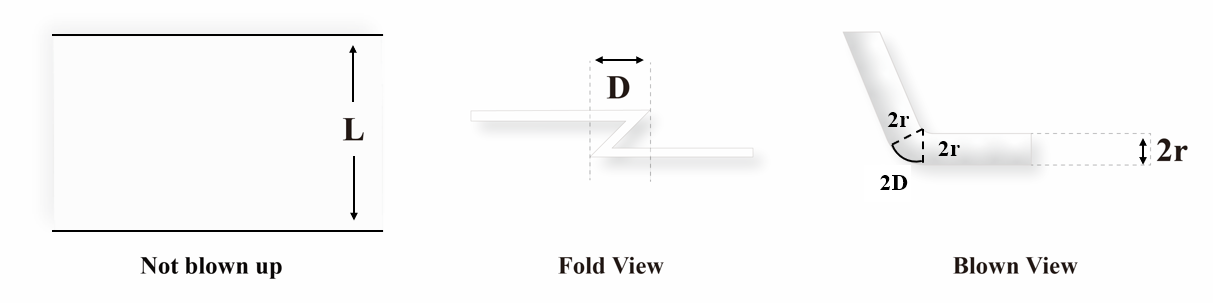}
    \caption{Wrinkle geometric parameters}
    \label{fig:tube}
\end{figure}
As shown in Fig.~\ref{fig:tube}, the basic parameters for the Vine robot include: $L$  width of the tube, $D$ length of the wrinkle, $\theta$ turning angle, $r$ radius of the tube after inflating.

The wrinkle formation process maintains a consistent fold length in the tube, as the wrinkle inducer arm and lower bonding arm share a single motor drive. This unified actuation ensures the arm rises to a predetermined height during each cycle. The design specifies a 15-degree rotation of the wrinkle inducer arm, which produces a precise 18mm fold in the tube.

Each wrinkle reduces the linear distance between two points by $2D$ on the compressed side. Since the PVC tube material has very little stretch, the length is conserved and the outer arc of the wrinkle maintains this $2D$ length. These geometric constraints allow us to derive the resulting rotation angle of the tube segment.

\begin{equation}
    \theta = \frac{D}{r} = \frac{D\pi}{L}
\end{equation}

In our case, $D = 19$mm and $L = 105$mm; thus, $\theta = 30.3^o$. In theory, this turning angle should be in 3D space. However, in this work, we only examine planar motion where the tube is moves on a surface. The net effect is equivalent to the projection of a turn in 3D space onto a flat surface.  Since our tape is strategically placed at the quarter point positions around the tube's circumference (forming a 45-degree angle with the horizontal plane), the actual angle on the table should be multiplied by $\cos(45^o)$. This quarter-point placement was chosen to balance the turning force and maintain structural stability during inflation. Therefore, the effective angle of turn is
$    \theta' = \theta\cos(45^\circ) \approx 21.44^\circ$.

\begin{figure*}[h]
    \centering
    \begin{subfigure}[b]{0.2\textwidth}
        \centering
        \includegraphics[width=\textwidth,angle=0]{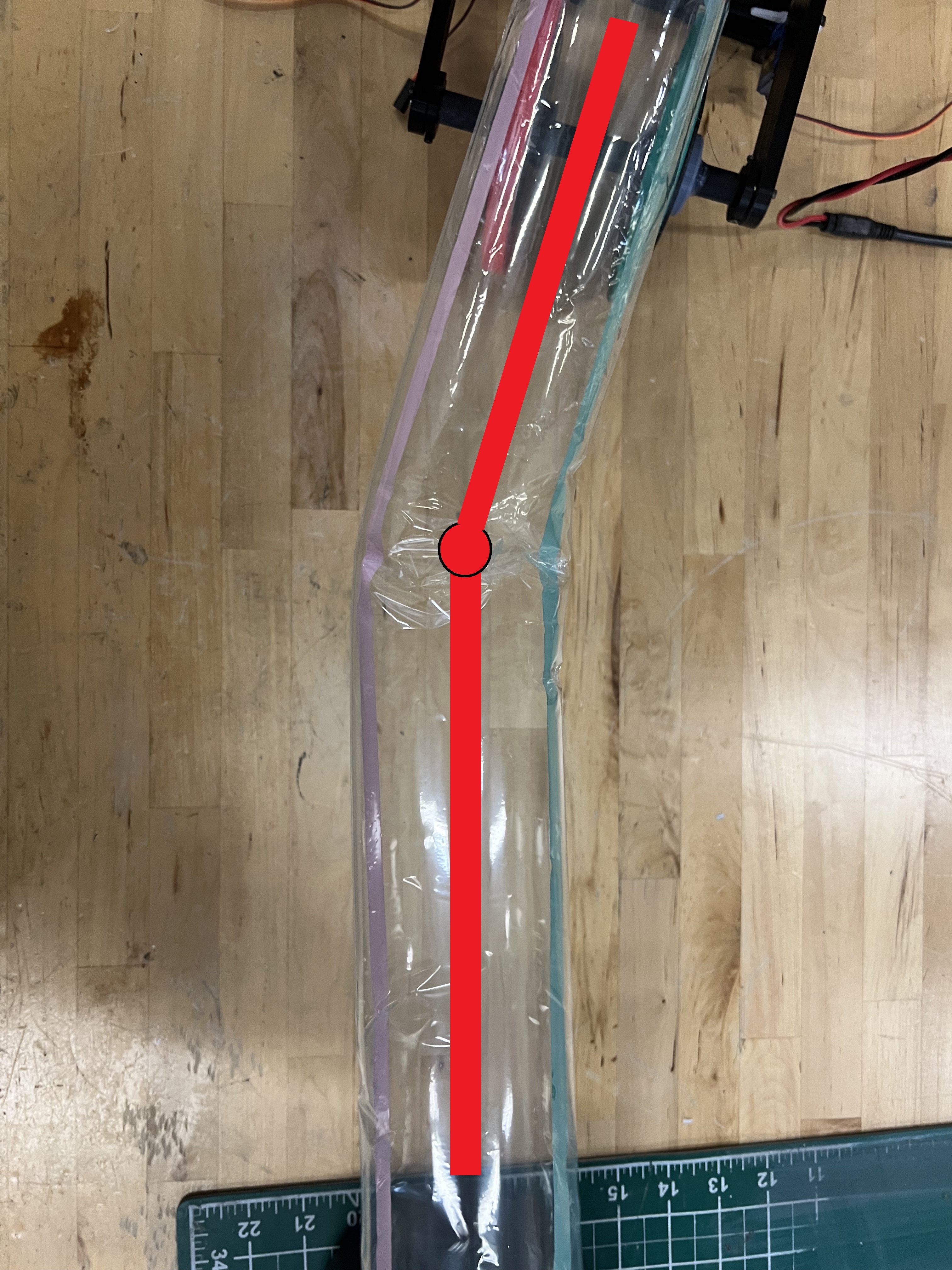}
        \caption{R}
    \end{subfigure}
    \hfill
    \begin{subfigure}[b]{0.235\textwidth}
        \centering
        \includegraphics[width=1.125\textwidth,angle=90]{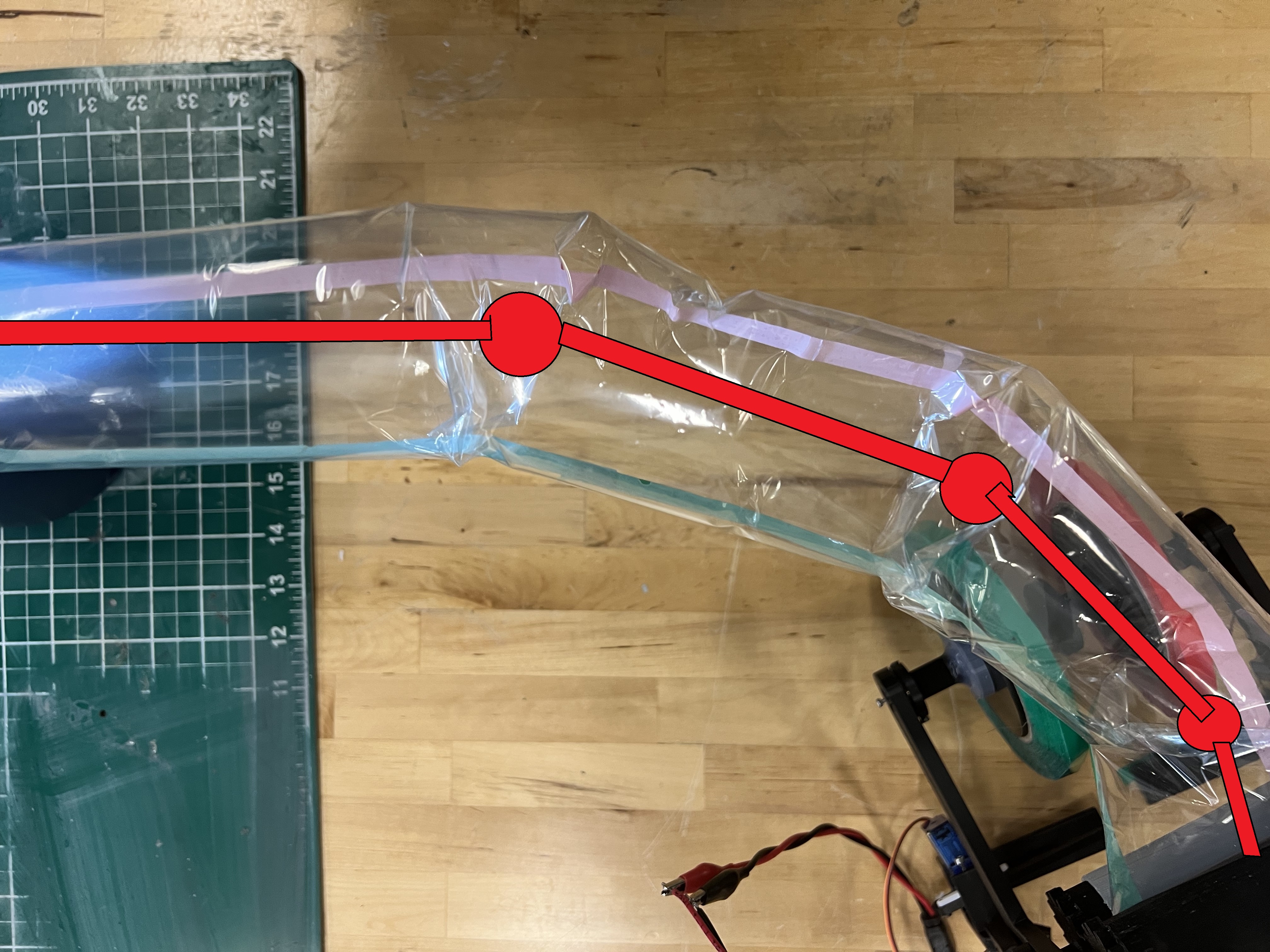}
        \caption{RRR}
    \end{subfigure}
    \hfill
    \begin{subfigure}[b]{0.2\textwidth}
        \centering
        \includegraphics[width=\textwidth,angle=0]{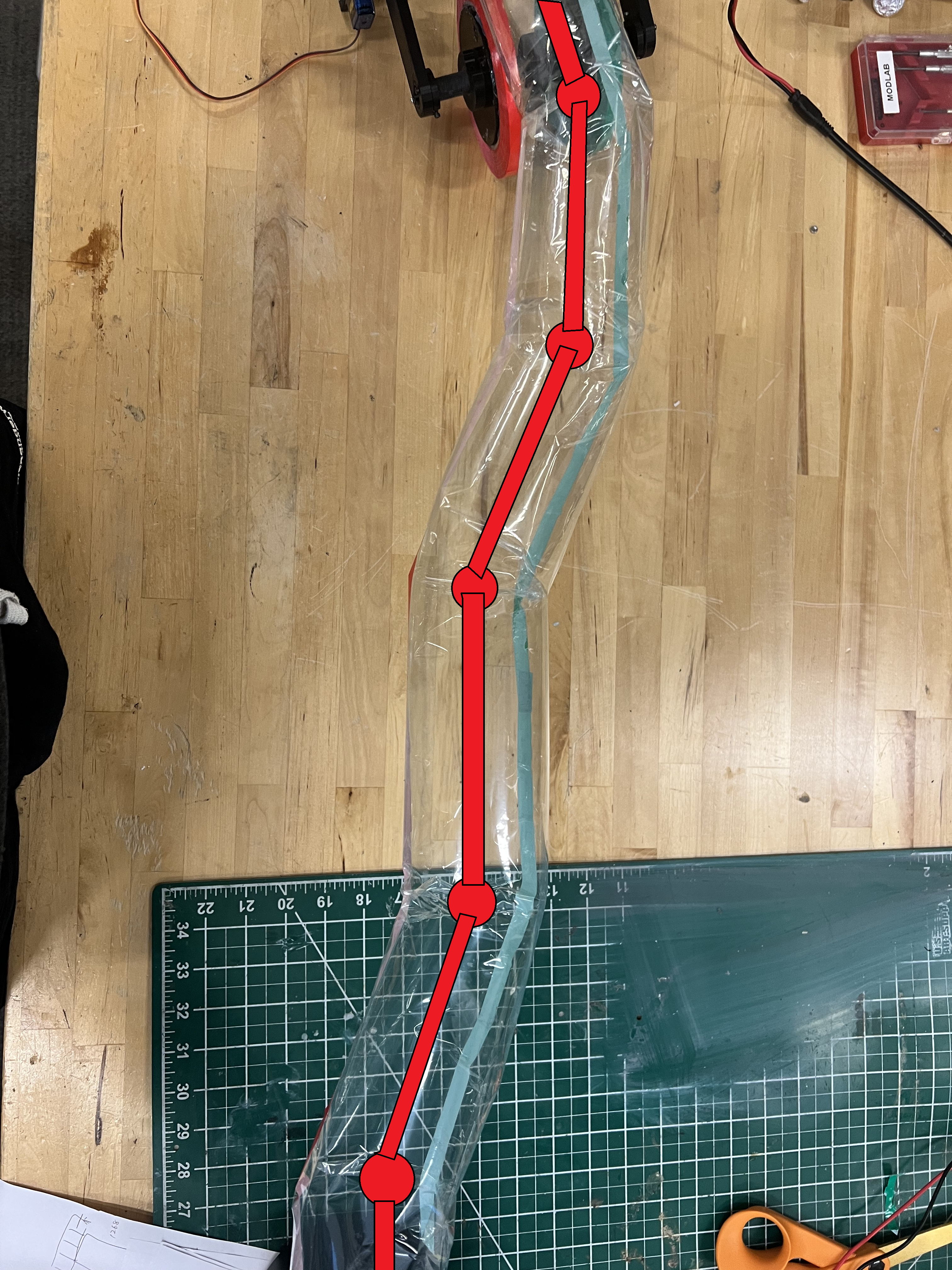}
        \caption{RLRLL}
    \end{subfigure}
    \caption{Experimental Validation of Multi-Turn Capabilities: (a) Single Right Turn (R) Demonstration, (b) Triple Consecutive Right Turns (RRR) Pattern, (c) Complex Turn Sequence (RLRLL) Demonstrating Bidirectional Control}
    \label{fig:Experiment}
\end{figure*}

\section{PATH PLANNING ALGORITHM}
\begin{figure}[H]
    \centering
    \includegraphics[width=0.5\linewidth]{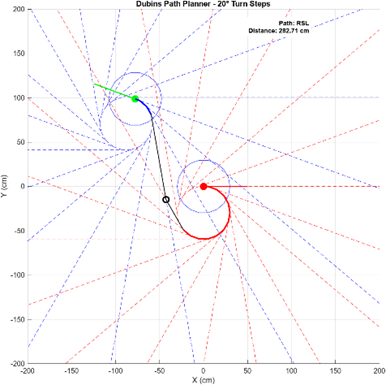}
    \caption{Segmented Dubins Path}
    \label{fig:dubins}
\end{figure}

\begin{algorithm}[h]
\caption{Generate Dubins Path }
\begin{algorithmic}
\Procedure{GenerateDubinsPath}{map, \textit{startPos}, \textit{startAngle}, \textit{endPos}, \textit{endAngle}, \textit{pathType}}
    \State map $\gets$ Precompute polygons and increments
    \State n $\gets$ 360/increment
    \For{$i \gets 1$ \textbf{to} $n$}
        \For{$j \gets 1$ \textbf{to} $n$}
            \State $intPt \gets$ \Call{Intersection}{map.startRays[i], map.endRays[j]}
            \If{$intPt$ exists}
                \State $angle \gets$ \Call{ComputeAngle}{map.startRays[i], map.endRays[j]}
                \If{$|angle - (180-increment)| < tolerance$}
                    \If{drawingEnabled} 
                        \State Draw the intersection on the map
                    \EndIf
                    \State $d \gets$ \Call{PathLength}{map, i, j, intPt}
                    \State \Return $d$, drawnHandles
                \EndIf
            \EndIf
        \EndFor
    \EndFor
    \State \Return NaN, null \Comment{No valid path found}
\EndProcedure
\end{algorithmic}
\label{alg:dubins}
\end{algorithm}

Conventional Dubins path planning \cite{dubins1957curves} computes the shortest feasible route by combining circular arcs with straight-line segments. While traditional Dubins paths assume continuous curvature control with any angle between segments, our Vine robot can only make discrete turns of approximately $21^o$ per wrinkle. This constraint changes the problem from continuous optimization to discrete geometric construction. Our modified algorithm discretizes the turning process into fixed-angle increments, approximating curves using sequences of short straight segments connected by $21^o$ turns.

In our implementation, the algorithm uses a ray-casting approach where \textit{map.startRays} and \textit{map.endRays} represent arrays of rays emanating from the start and end positions at different angles (discretized by \textit{increment}).The \textit{Intersection} function calculates where these rays intersect, creating potential transition points between the initial and final configurations. The \textit{ComputeAngle} function evaluates the angle formed at these intersections to ensure it satisfies our fixed-angle constraint of $21^o$ turns. For valid configurations, the algorithm computes path length using the \textit{PathLength} function and returns the distance and visualization handles. 

For example, a $63^o$ right turn would use three consecutive $21^o$ turns. For angles not exactly divisible by $21^o$ (e.g., 70 degrees), our algorithm implements a slight over-rotation (e.g., four $21^o$ turns for $84^o$) then applies a correction to guide the robot back to the intended path. This deviation-then-correction approach ensures alignment with the desired trajectory despite discrete turning constraints. While traditional Dubins paths achieve optimality through continuous control, our algorithm sacrifices some optimality to accommodate our Vine robot's physical constraints, producing paths that closely approximate optimal solutions with physically realizable $21^o$ turns.

\section{EXPERIMENTS }

Experimental evaluation consisted of three test scenarios to assess the system's steering stability and performance. In the first set, we measured the angular stability of the first turn; in the second set, we tested the angle of ten consecutive left or ten consecutive right turns; and in the third set, we tested the angular stability of the left and right intersections when making a wrinkle. 

\subsection{Results}
\begin{figure*}[h]  
    \centering
    \includegraphics[width=0.8\textwidth]{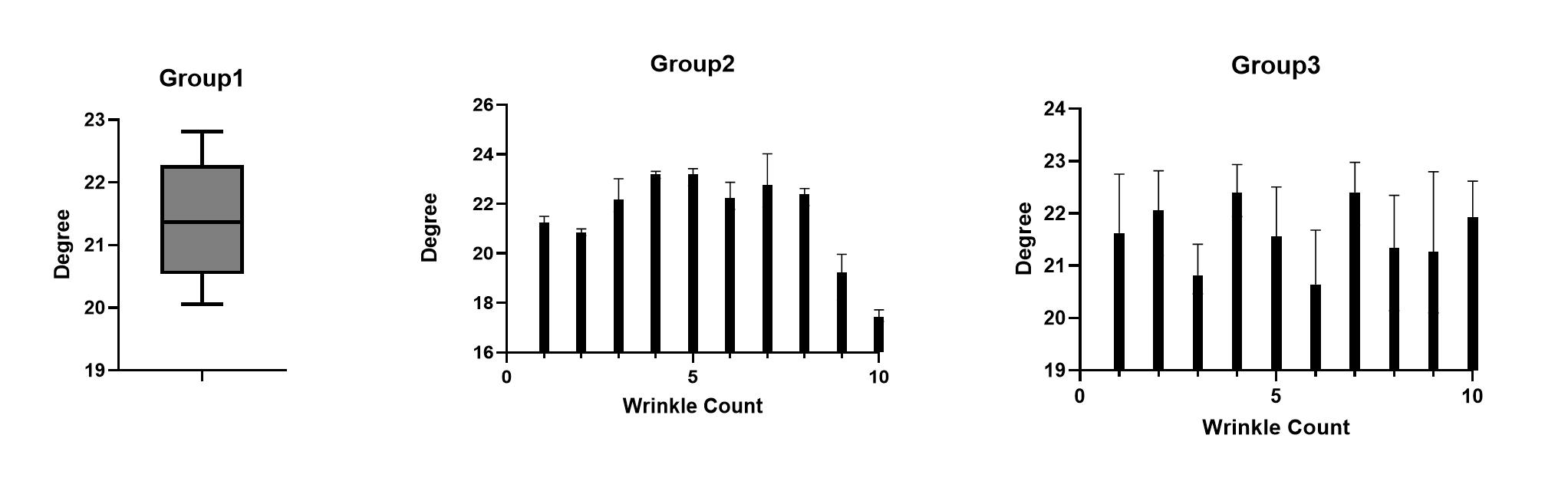}
    \caption{Performance Evaluation. Group 1 tests single turns. Group 2 tests sequential turns in one direction. Group 3 tests sequential alternating turns}
    \label{fig:Performance}
\end{figure*}

Results can be seen in Fig.~\ref{fig:Performance}.
In the first scenario, we evaluated single-turn stability under optimal conditions (no tube deflection or air leakage),  resetting between each turn. The system consistently achieved turning angles of 21.5  $\pm 1.5^o$.

The second scenario investigated consecutive unidirectional turns, either all left or all right. The first few turns had angles ranging from $20^o$ to $23^o$. However, we observed a systematic reduction in the bend angle as the number of wrinkles increased. Beyond ten consecutive turns, the accumulated material shifting ultimately led to system failure.

Alternating left-right turns in the third scenario assesses bidirectional steering stability. During the first ten turns, the system maintained consistent performance with an average turning angle of 21 $\pm 2^o$. The feeding material does not show progressive misalignment during operation, supporting the effectiveness of wrinkle formation. We expect the limit of the number of wrinkles to be up to the material supply.

\subsection{Error Analysis}

The single-turn stability tests demonstrated high consistency in steering angles, validating the feasibility of automating the wrinkle-based steering mechanism and the basic effectiveness of our mechanical design.

The progressive degradation from material shifting leads to two types of failures: First, the tape gradually deviates from its alignment, causing angular errors despite maintaining consistent wrinkle lengths. Second, the misalignment leads to air leakage issues, where air leaks through the rollers from the pressurized side to the unpressurized side. This introduces additional errors in the wrinkle formation process. 

Notably, the alternating left-right turn's improved performance suggests that the alternating helps neutralize the cumulative shifting effects.

\section{DISCUSSION}

\subsection{Challenges and Solutions}

This shifting problem is the main problem in this prototype. 

\paragraph{One Side Wrinkle Creation}

\begin{figure}[h]
    \centering
    \includegraphics[width=0.5\textwidth]{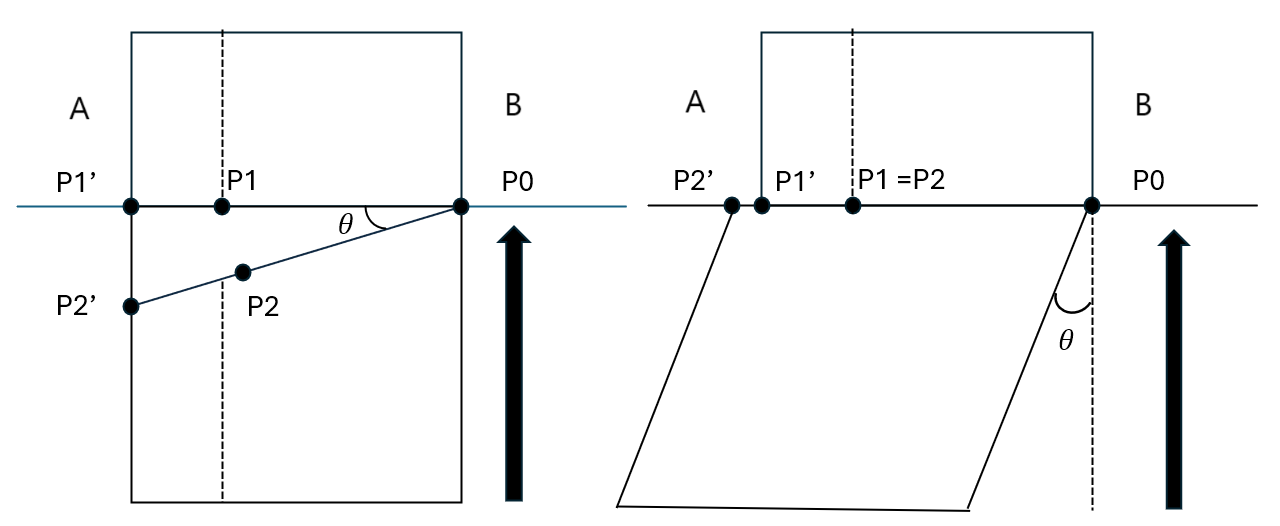}
    \caption{One Side Wrinkle geometry}
    \label{fig:wrinkle}
\end{figure}

When we generate a wrinkle, the feeding material will steer based on which side of the wrinkle is generated. In Fig.~\ref{fig:wrinkle}, the dashed line represents the tape, and the black line represents the tube shape. P1 and P2 are the points taped together where $$||P0P1|| = ||P0P2||$$ The feeding material will shift by $\theta$. Thus, the feeding material will lose alignment with the roller.

To maintain alignment, the two wrinkle inducer arms must lift together to create a rectangular wrinkle, making $\theta = 0$. However, a rectangular wrinkle is symmetric, so it will not turn the tube left or right. The compensation mechanism allows us to create the asymmetry by having allowing the tube to be symmetric before inflation then one side unfolds after inflation. Suppose that we are going to make a wrinkle on side A. Since in the rectangular-shaped wrinkle, side B is also folded.  To unfold side B the tape is adhered on the other side while connecting P1 and P2 directly. 
\paragraph{Feeding Material Line-up Error}

Ideally, the extended Vine robot has uninflated material from the source roll at the base to the end-device and back to the base on the inflated side of the air pump.  Here, the uninflated material should match the inflated section, including turns. However, when the length of the feeding material is too long, or the resistance of the tabletop is too big to prevent the main body of the Vine robot from turning, there is a force between the roller and the tube, which causes the tube to veer from the intended direction. 

\paragraph{Air Leakage Problem}

Leakage is another crucial issue in this project. Since we have the feeding material external to the Vine robot, air can leak from the inflated side to the non-inflatable side. This mainly occurs during the wrinkle transfer, where the thickness of the material passing the rollers is increased from the wrinkle. The rubber layer in the middle of the rollers and suitable compression between the rollers prevent leakage. However, increased compression requires greater torque from the motors driving the rollers.

\subsection{Limitations}

This Vine robot performs real-time control,  length, and turning up to material limits and maintains a soft structure compared to other Vine robots. However,  this Vine robot can only turn at a fixed angle each time. For example, the structure in this paper can only turn at a fixed angle of $21^o$ to the left or right each time. This problem can be solved by adding mechanisms to change the length of the wrinkle inducer. For example, adding two servos to control the folding length of each wrinkle can enable arbitrary angle control. 

In addition, the current prototype is limited to planar motion partly due to the heavy end-device and the lack of stability of the feed material supply.

Finally, the feeding material's stability is currently uncontrolled and may prevent accurate motion. A simple, lightweight structure must keep the feeding material parallel to the roller and stabilize it in the center, avoiding side-to-side deviation.

\section{CONCLUSION AND FUTURE WORK}
This paper presents a novel approach to real-time steering control for Vine robots through surface wrinkle fabrication. Our implementation successfully demonstrates real-time planar steering control using a dual-tape wrinkle generation mechanism while maintaining the robot's soft body structure and preserving its characteristic features of theoretically infinite length and turning capabilities. Experimental results validate the effectiveness of this approach, showing consistent $21^o$ turns with error margins within $2^o$ for initial operation, though performance degradation was observed after multiple consecutive turns.

Our future work will focus on extending the system's capabilities to 3D space operation. This requires a more sophisticated mathematical model for 3D wrinkle formation and creating adaptive path planning algorithms for three-dimensional navigation. We will address the current mechanical limitations by reducing the end-device mass through materials optimization and component redesign. Particular attention will be paid to resolving the material shifting problem through new feeding mechanisms and developing a more robust pressure control system for improved stability reducing the roller motor torque requirements allowing for smaller lighter end-device.

\section*{ACKNOWLEDGMENT}

We want to thank Gregory Campbell for his guidance  on the challenges. Additionally, we thank everyone who contributed to this project, including Claudia Liu, Derek Ike, and Tej Panigrahi.

\bibliographystyle{IEEEtran}
\bibliography{bibliography}

\end{document}